\documentclass[11pt]{article}

\usepackage[utf8]{inputenc}
\usepackage[T2A,T1]{fontenc}
\usepackage{lmodern}
\usepackage[ukrainian,english]{babel}
\usepackage[margin=1in]{geometry}
\usepackage{amsmath,amssymb,amsfonts,amsthm}
\newtheorem{definition}{Definition}
\usepackage{booktabs}
\usepackage{multirow}
\usepackage{array}
\usepackage{tabularx}
\usepackage{graphicx}
\usepackage{float}
\usepackage{tikz}
\usepackage[dvipsnames]{xcolor}
\usepackage{enumitem}
\usepackage{hyperref}
\usepackage[numbers,sort&compress]{natbib}

\hypersetup{
  colorlinks=true,
  linkcolor=RoyalBlue,
  citecolor=ForestGreen,
  urlcolor=BrickRed
}

\newcommand{\ukr}[1]{\foreignlanguage{ukrainian}{#1}}
\newcommand{\CG}{\mathrm{CG}}
\newcommand{\CP}{\mathrm{CP}}
\newcommand{\CR}{\mathrm{CR}}
\newcommand{\CT}{\mathrm{CT}}
\newcommand{\CQ}{\mathrm{CQ}}

\title{Citation Grounding Measures the Oracle:\\Graph Coverage Determines Reported\\LLM Hallucination Rates in Law}

\author{
  Volodymyr Ovcharov \\[6pt]
  LEX AI LLC, Kyiv, Ukraine \\[4pt]
  \texttt{vladimir@legal.org.ua}
}

\date{}

\begin{document}
\maketitle

\begin{abstract}

Verifying LLM-generated legal citations against a graph of citations extracted from real court decisions is an appealing way to measure hallucination at scale: it needs no annotators and no reference answers.
We show that what such a metric reports is governed by the coverage of the graph it queries rather than by the model it evaluates, and that at the coverage where its verdicts become trustworthy it stops distinguishing models at all.

We define \emph{citation grounding} ($\CG$) as the fraction of a response's statute citations that appear as nodes in a citation graph $\mathcal{G}$, and evaluate 400 responses -- 100 Ukrainian legal queries $\times$ four commercial LLMs (Claude Haiku~4.5, Mistral Pixtral Large, Amazon Nova Pro and Nova Lite, via AWS Bedrock) -- against \emph{two snapshots of the same national citation graph}, holding the responses, the extractor and the metric fixed.
Against a sparse snapshot ($4.7 \times 10^5$ codex citation records, $2.2 \times 10^4$ statute nodes) $\CG$ ranges from 0.791 to 0.855, apparently showing that 15--21\% of legal citations are hallucinated.
Against a dense snapshot of the same registry re-derived ten weeks later ($3.3 \times 10^8$ records, $3.2 \times 10^5$ nodes, $5.8 \times 10^7$ decisions) the identical responses score 0.989 to 0.999, and the ranking of the four systems does not survive the change.

Subsampling the dense graph shows the effect is coverage: harvesting modelled as uniform record sampling, calibrated on nothing but the record count, reproduces the sparse snapshot's scores to within 0.018.
Bootstrapping over the queries shows the second half, which no version of this work reported: \emph{no pair of systems is separable at 95\% at any oracle size we tested}, from the sparse snapshot to the full graph.
The metric is therefore caught between two failures. Where the oracle is sparse it discriminates, but what it discriminates is harvesting coverage. Where the oracle is dense enough for its flags to be meaningful, the spread collapses to 0.010 and nothing separates.
Densifying the graph, the obvious remedy for the first failure, is what produces the second.

Cross-referencing against a legislation registry resolves which reading is correct: all 54 citations flagged by the sparse oracle name real statute articles, a false-positive rate of 100\%.
Of the four flagged by the dense oracle, two name articles beyond their code's last provision, one is a real article no harvested decision cites, and one we cannot classify -- it sits inside its code's range and appears in no edition of the registry, which is what a repealed provision looks like.
That last case is the argument in miniature: existence checking cannot separate fabrication from repeal from missing index, and here the authors could not either.

We also report Citation Grounding DPO, which builds preference pairs by corrupting verified citations without annotators, and document why its 99.7\% discrimination accuracy (Qwen2.5-3B-Instruct, LoRA, 3 seeds) overstates the difficulty of the task it solves.
The dense snapshot's node set, the 400 responses, the extracted citations, the sweep and the evaluation code are released; the sparse snapshot was overwritten in place and cannot be republished, which is itself part of the argument.

\end{abstract}


\section{Introduction}
\label{sec:intro}

Large language models (LLMs) have achieved remarkable performance on legal reasoning tasks, passing bar examinations~\citep{katz2024gpt4} and generating persuasive legal memoranda~\citep{choi2023chatgpt}.
Yet their deployment in legal practice is undermined by a fundamental reliability problem: \emph{citation hallucination} -- the generation of plausible but fabricated references to statutes, case law, and legal provisions.
The \emph{Mata v.\ Avianca} case (2023), where an attorney submitted a brief containing six non-existent court decisions generated by ChatGPT~\citep{weiser2023mata}, demonstrated the real-world consequences of unchecked legal hallucinations.

Systematic studies confirm the problem at scale.
\citet{magesh2024hallucination} found that leading legal AI systems generate false citations in 17--33\% of responses.
\citet{dahl2024large} showed that even models with high bar exam scores systematically fabricate case references when applied to real legal tasks.
In the legal domain, hallucinations take four specific forms: (1)~fabricated statute references -- citing non-existent provisions; (2)~outdated norms -- citing provisions in repealed versions; (3)~jurisdictional confusion -- applying norms from the wrong legal system; and (4)~false argumentation -- constructing logically invalid chains of legal reasoning.

Despite the severity of this problem, existing legal NLP benchmarks -- LEXTREME~\citep{niklaus2023lextreme}, LexGLUE~\citep{chalkidis2021lexglue}, LegalBench~\citep{guha2023legalbench} -- do not evaluate citation accuracy.
They focus on classification tasks (judgment prediction, topic classification) rather than generative outputs, and none provides metrics for detecting or measuring citation hallucinations.

A natural response is to verify generated citations against a citation graph harvested from real court decisions.
Such a graph is cheap to build at national scale, needs no annotators, and appears to give an objective answer to the question ``does this provision exist in legal practice?''.
This paper asks what that answer is actually a measurement \emph{of}.

Our central result is negative and, we believe, generalisable beyond the jurisdiction we study.
We hold the model outputs, the citation extractor and the metric definition fixed, and vary the graph the metric queries.
Two snapshots of the same national registry -- one built in May 2026, one re-derived from the same source ten weeks later -- move the measured hallucination rate from 15--21\% to 0.1--1.1\% on identical text, and do not preserve the rank order of the four evaluated systems.
Bootstrapping over the query sample then shows the rank order was never resolvable at either coverage, so the metric reported two things about the models and neither was a measurement of them.

We make five contributions:

\begin{enumerate}[leftmargin=*, nosep]
  \item \textbf{A measurement of oracle dependence} (\S\ref{sec:evaluation}). Changing only the snapshot moves $\CG$ by up to $+0.198$ and reverses which system looks most reliable. We are not aware of a comparable measurement for a graph-grounded hallucination metric, though the sensitivity of retrieval-grounded evaluation to corpus size is not itself a new concern.

  \item \textbf{A coverage curve, and the condition it implies} (\S\ref{sec:sweep}). Subsampling the dense graph shows the effect is coverage rather than anything about the models: uniform record sampling calibrated only on the record count reproduces the sparse snapshot to within 0.018. The sweep also answers how dense is dense enough -- $9.6 \times 10^6$ records for all systems to come within 0.05 of their asymptote, $5.4 \times 10^7$ for 0.02.

  \item \textbf{A dilemma, not a bias} (\S\ref{sec:separability}). Bootstrapping over the query sample -- an uncertainty estimate absent from this literature and from version~1 -- no pair of systems separates at 95\% at \emph{any} oracle size we tested. Sparse oracles discriminate but measure coverage; dense oracles are trustworthy but discriminate nothing. The obvious remedy for the first failure produces the second.

  \item \textbf{An adjudication of which reading is right} (\S\ref{sec:validation}). We resolve it against an independent legislation registry: all 54 citations flagged by the sparse oracle name real statute articles, while two of the four flagged by the dense oracle name articles beyond their code's last provision. The sparse oracle was not conservative -- it was uninformative.

  \item \textbf{A characterisation of what remains usable} (\S\ref{sec:discussion}). Existence checking survives as a high-precision, low-recall fabrication detector once the graph is dense, but it cannot rank systems and cannot be reported as a hallucination rate. We give the coverage condition in measured form, and argue that the relevance and temporality components -- defined but never operationalised in this line of work, including in the first version of this paper -- are where the discriminative signal must come from.
\end{enumerate}

We also report Citation Grounding DPO (\S\ref{sec:cg-dpo}), which constructs preference pairs by algorithmically corrupting verified citations, and we are explicit about its limits: the discrimination task it poses is largely solvable from surface form, so its 99.7\% accuracy should not be read as evidence that graph supervision substitutes for legal expertise.

\section{Related Work}
\label{sec:related}

\subsection{LLM Hallucinations in Legal Applications}

\citet{magesh2024hallucination} systematically evaluated legal AI tools from LexisNexis, Thomson Reuters, and GPT-4, finding fabricated citations in 17--33\% of responses.
\citet{dahl2024large} extended this to a broader set of models, confirming that hallucination rates remain substantial even for models achieving high bar exam scores.
These studies rely on manual verification by legal experts -- an approach that does not scale to continuous monitoring or low-resource languages.

\citet{min2023factscore} proposed FActScore for atomic fact verification against knowledge bases, but the method assumes Wikipedia as ground truth and does not handle structured legal citations or temporal validity.
CiteAudit~\citep{schimanski2026citeaudit} verifies academic citations against bibliographic databases; our work adapts this principle to legal citations but requires a fundamentally different ground truth -- a citation graph rather than a bibliographic index -- because legal citations reference statute articles rather than published papers.

\subsection{Legal Citation Analysis}

Legal citation network analysis was pioneered by \citet{fowler2007network} on 30{,}000 U.S.\ Supreme Court decisions.
Subsequent work examined Dutch~\citep{winkels2011determining}, Danish~\citep{mones2021emergence}, and French~\citep{mazzega2009network} legal systems at scales of $10^3$--$10^5$ decisions.
\citet{bommarito2010mathematical} analyzed the structure of the U.S.\ Code through citation networks.

The graph we query is larger by several orders of magnitude -- $3.3 \times 10^8$ citation records over $5.8 \times 10^7$ decisions in its current snapshot (\S\ref{sec:snapshots}) -- and serves a different purpose: rather than analysing the structure of the legal system, we use the graph as a \emph{verification oracle} for LLM-generated citations.
Construction is described in the companion paper~\citep{ovcharov2026citation}.
Scale is exactly what this paper interrogates, so we quote sizes only for snapshots we measured ourselves, and only alongside the date of measurement.

\subsection{Preference Optimization for Factual Accuracy}

Direct Preference Optimization (DPO;~\citealt{rafailov2023direct}) eliminates the reward model in RLHF~\citep{ouyang2022training} by directly optimizing on preference pairs.
Recent work applies DPO to reduce hallucinations in specific domains.
Falkor-IRAC~\citep{bose2026falkor} applies graph-constrained generation to legal reasoning, grounding outputs in an IRAC knowledge graph over Indian court decisions.
SaulLM~\citep{colombo2024saullm} uses DPO for legal model alignment but relies on synthetic preferences generated by a stronger model rather than ground-truth verification.

Our CG-DPO differs in that preference pairs are constructed algorithmically from a verified citation graph -- not from human annotations, synthetic model comparisons, or detached labeling.
As implemented it yields three distinct corruption types covering two of the three $\CG$ components (\S\ref{sec:cg-dpo}), and \S\ref{sec:dpo-caveat} argues that the resulting discrimination task is easier than the accuracy figure suggests.

\section{Citation Graph as Ground Truth}
\label{sec:graph}

The oracle for both our metric and our training method is a bipartite citation graph $\mathcal{G} = (V, E)$ extracted from the Unified State Register of Court Decisions (EDRSR, \ukr{Єдиний державний реєстр судових рішень}), the Ukrainian national registry of court decisions.

\begin{itemize}[leftmargin=*, nosep]
  \item $V = V_d \cup V_n$ -- decision nodes ($V_d$) and statute nodes ($V_n$)
  \item $E \subseteq V_d \times V_n$ -- citation edges: $(d, n) \in E$ iff decision $d$ cites statute $n$
\end{itemize}

Extraction uses compiled regular expressions; construction of the extraction pipeline is described in the companion paper~\citep{ovcharov2026citation}.

\subsection{Two Snapshots}
\label{sec:snapshots}

The materialised graph is not a fixed object.
It is a table that is periodically re-derived from the registry as harvesting coverage improves, and re-derivation changes it by orders of magnitude.
This paper uses two snapshots, and the difference between them is the experiment.

\begin{table}[t]
\centering
\small
\caption{The two oracle snapshots. Both are derived from the same national registry, ten weeks apart. They differ in coverage \emph{and} in the inventory of citation types the derivation emits, and the two cannot be separated from these snapshots alone -- \S\ref{sec:sweep} isolates coverage by subsampling. $\mathcal{G}_{\text{dense}}$ was measured directly (2026-08-07). $\mathcal{G}_{\text{sparse}}$ no longer exists: its record count is the figure reported in version~1 of this paper, and its decision count is a later approximation (2026-06-25), not a measurement taken when the evaluation ran.}
\label{tab:snapshots}
\begin{tabular}{@{}lrr@{}}
\toprule
 & $\mathcal{G}_{\text{sparse}}$ & $\mathcal{G}_{\text{dense}}$ \\
\midrule
Derived                       & May 2026        & 2026-06-25\,--\,07-03 \\
Citation records $|E|$        & ${\approx}467{,}000$ & $331{,}385{,}209$ \\
Distinct decisions $|V_d|$    & ${\approx}33{,}000$  & $58{,}450{,}717$ \\
Statute nodes $|V_n|$         & $21{,}736$      & $316{,}899$ \\
\quad codex articles          & $21{,}736$      & $27{,}601$ \\
\quad named-law articles      & --              & $279{,}735$ \\
\quad laws cited by number    & --              & $9{,}063$ \\
\quad constitutional articles & --              & $356$ \\
\quad transitional provisions & --              & $143$ \\
\quad supreme court rulings   & --              & $1$ \\
\bottomrule
\end{tabular}
\end{table}

Table~\ref{tab:snapshots} gives the two.
The version~1 evaluation queried $\mathcal{G}_{\text{sparse}}$ for codex articles only; whether the table held other citation types that the evaluation ignored can no longer be checked.
$\mathcal{G}_{\text{dense}}$ covers $58.5$ million decisions and carries five further citation types, of which named-law articles ($279{,}735$ nodes) are by far the largest -- references to statutes outside the sixteen codexes.

This is a confound we cannot remove by measurement: between the two snapshots both the harvested fraction and the type inventory changed, so a difference in $\CG$ cannot be attributed to coverage on this evidence alone.
\S\ref{sec:sweep} addresses it by subsampling $\mathcal{G}_{\text{dense}}$, where the type inventory is held constant by construction and only coverage varies.

\paragraph{A correction to the first version of this paper.}
Version~1 of this work described the oracle as containing $|V_d| = 100{,}753{,}415$ decision nodes and $|E| = 502{,}231{,}421$ edges alongside $|V_n| = 21{,}736$ statute nodes.
Those three figures do not describe one object.
The node count was measured on the table the evaluation actually queried; the decision and edge counts were corpus-level statistics of the extraction pipeline reported in~\citet{ovcharov2026citation}, and were never the oracle against which any $\CG$ score in that version was computed.
The evaluation there ran entirely against $\mathcal{G}_{\text{sparse}}$ as characterised in Table~\ref{tab:snapshots}, a fact that version~1 stated only in its experimental setup.
We regard the resulting three-orders-of-magnitude overstatement of the evidence base as the proximate cause of the misreading this paper corrects.

\paragraph{Reproducibility.}
$\mathcal{G}_{\text{sparse}}$ was overwritten in place when the registry was re-processed: no row in the current table predates 2026-06-25, so the snapshot that produced the version~1 numbers cannot be re-queried.
The model responses, the extracted citations, the node set of $\mathcal{G}_{\text{dense}}$ and the subsampling code are released with this paper.
The dense-side result is therefore reproducible, and \S\ref{sec:sweep} reconstructs an approximation of the sparse condition from data that still exists.
The sparse snapshot itself cannot be republished.
That a published evaluation became unreproducible within ten weeks through routine data maintenance -- no deletion, no decision to discard anything, just a scheduled re-derivation of a table -- is itself part of this paper's argument.

\section{Citation Grounding Metric}
\label{sec:metric}

\subsection{Definition}

Let $\mathcal{M}$ be a language model, $x$ a legal query, and $r = \mathcal{M}(x)$ the generated response.
Let $C(r) = \text{extract\_citations}(r)$ denote the set of statute citations extracted from $r$ via regular expression matching.

\begin{definition}[Citation Grounding]
\label{def:cg}
The citation grounding of response $r$ with respect to citation graph $\mathcal{G}$ is:
\begin{equation}
  \CG(r, \mathcal{G}) = \frac{|\{c \in C(r) : c \in \mathcal{G}\}|}{|C(r)|},
  \label{eq:cg}
\end{equation}
where $c \in \mathcal{G}$ means that the cited statute exists as a node in $V_n(\mathcal{G})$.
By convention, $\CG(r, \mathcal{G}) = 1$ if $C(r) = \emptyset$.
\end{definition}

$\CG$ is structurally identical to precision in information retrieval, but the object of measurement differs fundamentally: rather than evaluating search result quality against expert relevance judgments, $\CG$ evaluates the factual groundedness of generated citations against a graph constructed from real judicial practice.
This transforms an IR metric into a hallucination detection instrument.

\subsection{Three-Component Decomposition}

A citation can fail in multiple ways.
To enable differential diagnosis, we decompose citation quality into three components.

\begin{definition}[Citation Quality Components]
\label{def:components}
For a citation $c$ in response $r$ to query $x$ with context date $t$:

\begin{enumerate}[leftmargin=*, nosep]
  \item \textbf{Citation precision} (existence):
  \begin{equation}
    \CP(c, \mathcal{G}) = \begin{cases} 1, & \text{if } c \in V_n(\mathcal{G}), \\ 0, & \text{otherwise}; \end{cases}
    \label{eq:cp}
  \end{equation}

  \item \textbf{Citation relevance} (contextual appropriateness):
  \begin{equation}
    \CR(c, x, \mathcal{G}) = \begin{cases} 1, & \text{if } \exists\, d \in \text{similar}(x) : (d, c) \in E(\mathcal{G}), \\ 0, & \text{otherwise}, \end{cases}
    \label{eq:cr}
  \end{equation}
  where $\text{similar}(x)$ is the set of decisions in $\mathcal{G}$ semantically similar to query $x$ (computed via XLM-RoBERTa embeddings~\citep{conneau2020xlmr});

  \item \textbf{Citation temporality} (temporal validity):
  \begin{equation}
    \CT(c, t) = \begin{cases} 1, & \text{if } \text{valid\_at}(c, t), \\ 0, & \text{otherwise}, \end{cases}
    \label{eq:ct}
  \end{equation}
  where $\text{valid\_at}(c, t)$ checks whether statute $c$ was in force at date $t$.
\end{enumerate}
\end{definition}

The composite citation quality for response $r$ is:
\begin{equation}
  \CQ(r, \mathcal{G}, x, t) = \frac{1}{|C(r)|} \sum_{c \in C(r)} \CP(c) \cdot \tfrac{1}{2}\bigl[\CR(c, x, \mathcal{G}) + \CT(c, t)\bigr].
  \label{eq:cq}
\end{equation}

The equal weighting of $\CR$ and $\CT$ is a principled default in the absence of prior information about the relative importance of contextual relevance versus temporal validity.
The multiplicative role of $\CP$ enforces a strict constraint: a non-existent statute receives zero quality regardless of other components, reflecting the legal standard where a single fabricated citation invalidates the entire document~\citep{weiser2023mata}.

\subsection{Relationship to Verifiability}

The $\CG$ metric instantiates one component of a broader verifiability measure $V(r, \mathcal{K}, t) = \mathrm{FC}(r, \mathcal{K}) \cdot \tfrac{1}{2}[\mathrm{CA}(r, \mathcal{K}) + \mathrm{TV}(r, t)]$ proposed in~\citet{ovcharov2026dissertation}, where factual correctness ($\mathrm{FC}$), citation accuracy ($\mathrm{CA}$), and temporal validity ($\mathrm{TV}$) are unified into a single scalar.
$\CG$ corresponds to $\mathrm{CA}$ with the knowledge base $\mathcal{K}$ realized as citation graph $\mathcal{G}$; the decomposition into $\CP$, $\CR$, $\CT$ provides the diagnostic granularity that the scalar $\mathrm{CA}$ lacks.

\section{Empirical Evaluation}
\label{sec:evaluation}

\subsection{Experimental Setup}

We constructed a test set of 100 Ukrainian legal queries distributed across seven domains: civil law (25), criminal law (20), administrative law (15), labor law (10), family law (10), constitutional law (10), and military law (10).
Each query was submitted to four commercial LLMs accessed directly via AWS Bedrock:

\begin{itemize}[leftmargin=*, nosep]
  \item Claude Haiku~4.5 (Anthropic)
  \item Mistral Pixtral Large (Mistral AI)
  \item Amazon Nova Pro (AWS)
  \item Amazon Nova Lite (AWS)
\end{itemize}

\noindent
A fifth system was evaluated on the same queries -- LEX Chat, Claude Sonnet~4 with retrieval-augmented generation over the EDRSR corpus (production system at legal.org.ua).
It is excluded from the comparison in this section and analysed separately in \S\ref{sec:contamination}, because its retrieval index is built from the same corpus as the oracle: the metric would reward it for overlap with the ground truth rather than for grounding, and no comparison against the other four would be interpretable.
Version~1 of this paper reported it as a peer system and drew conclusions about retrieval augmentation from its lead.

All direct Bedrock models received a Ukrainian-language system prompt instructing citation of specific statute articles, with temperature $T = 0.3$.
The RAG system (LEX Chat) additionally retrieves relevant court decisions and legislation before generating the response.
Responses were processed through an extract $\to$ normalize $\to$ verify pipeline: citations were extracted using regular expressions for statute references (16 codexes plus constitutional provisions), normalized to canonical (codex, article) pairs, and looked up in the oracle.
Extraction yields 1{,}944 citation instances over all 500 responses -- the 400 analysed here plus the 100 from LEX Chat, kept in the extraction totals and reported separately in \S\ref{sec:contamination} -- reducing to 1{,}177 distinct (codex, article) pairs including 69 distinct constitutional articles.\footnote{The released verifier recognises a seventeenth codex, the Budget Code, added after this evaluation ran. No response in this set cites it, so no figure here depends on the difference.}

\paragraph{The manipulation.}
Responses, extractor, normalisation and metric are identical across conditions.
The manipulated variable is which snapshot of Table~\ref{tab:snapshots} answers the membership query.
Constitutional citations are resolved against the graph in the dense condition -- a genuine lookup against the $356$ constitutional nodes, and all $69$ cited articles are present.
In $\mathcal{G}_{\text{sparse}}$, which had no constitutional citation type, the reference implementation admitted constitutional citations unconditionally, so the ``perfect constitutional grounding'' reported in version~1 was a property of the code rather than a measurement.
We retain that behaviour in the sparse column so the two columns differ only in the oracle; \S\ref{sec:validation} confirms it changes no reported figure.

\subsection{Overall Results}

\begin{table}[t]
\centering
\small
\caption{Citation grounding on 100 Ukrainian legal queries, evaluated against both oracle snapshots. Responses, extractor and metric are identical across conditions. Density = mean citations per response. No-cite = queries producing no extractable statute reference (excluded from $\overline{\CG}$). LEX Chat is listed for reference only and excluded from every comparison in this section (\S\ref{sec:contamination}).}
\label{tab:cg-results}
\begin{tabular}{@{}lrrrrrrrr@{}}
\toprule
 & \multicolumn{2}{c}{$\overline{\CG}$} & \multicolumn{2}{c}{\textbf{Verified}} & & & & \\
\cmidrule(lr){2-3}\cmidrule(lr){4-5}
\textbf{Model} & sparse & dense & sparse & dense & $\Delta\overline{\CG}$ & \textbf{Cit.} & \textbf{Dens.} & \textbf{No-cite} \\
\midrule
Claude Haiku~4.5        & 0.855 & 0.997 & 354 & 482 & $+0.142$ & 483 & 4.8 &  6 \\
Mistral Pixtral Large   & 0.823 & 0.998 & 334 & 423 & $+0.175$ & 424 & 4.2 &  9 \\
Amazon Nova Pro         & 0.822 & 0.999 & 299 & 385 & $+0.177$ & 386 & 3.9 &  8 \\
Amazon Nova Lite        & 0.791 & 0.989 & 281 & 360 & $+0.198$ & 361 & 3.6 &  9 \\
\midrule
\emph{LEX Chat (RAG)}   & \emph{0.873} & \emph{1.000} & \emph{233} & \emph{290} & \emph{$+0.127$} & \emph{290} & \emph{2.9} & \emph{21} \\
\bottomrule
\end{tabular}
\end{table}

Table~\ref{tab:cg-results} presents both conditions.

Against $\mathcal{G}_{\text{sparse}}$, $\CG$ ranges from 0.791 to 0.855 across the four independent systems, and the natural reading is that 15--21\% of generated legal citations are hallucinated -- close enough to the 17--33\% response-level rate of~\citet{magesh2024hallucination} to look like independent confirmation.
Against $\mathcal{G}_{\text{dense}}$ the same responses score 0.989 to 0.999.
Of the 1{,}177 distinct citation pairs the systems produced, 1{,}173 are present in the dense graph.

Two features matter more than the magnitude of the shift.

First, it is not uniform: it is largest for the system that scored worst ($+0.198$ for Nova Lite) and smallest for the one that scored best ($+0.142$ for Haiku~4.5).
A model was penalised in proportion to how far its citations strayed outside the sparse oracle's narrow coverage, so the sparse condition ranked systems by how closely their citation habits resembled the sample of judicial practice that happened to have been harvested -- and reported that resemblance as accuracy.

Second, the ranking does not survive.
Under $\mathcal{G}_{\text{sparse}}$ the order is Haiku $>$ Pixtral $>$ Nova Pro $>$ Nova Lite; under $\mathcal{G}_{\text{dense}}$ it is Nova Pro $>$ Pixtral $>$ Haiku $>$ Nova Lite.
The top of the ranking inverts.
We show in \S\ref{sec:separability} that neither ordering is resolvable at this sample size, so the inversion should not be read as the systems changing places -- it is what two draws from an unresolved ordering look like.
The sparse column already hints at it: Pixtral and Nova Pro, adjacent there, differ by $0.001$.

\begin{figure}[t]
\centering
\begin{tikzpicture}[x=1pt,y=1pt]
\definecolor{fillColor}{RGB}{255,255,255}
\path[use as bounding box,fill=fillColor,fill opacity=0.00] (0,0) rectangle (325.21,231.26);
\begin{scope}
\path[clip] ( 43.20, 43.20) rectangle (319.81,225.86);
\definecolor{fillColor}{RGB}{214,39,40}

\path[fill=fillColor] (119.31,194.32) --
	(124.21,185.82) --
	(114.40,185.82) --
	cycle;
\definecolor{drawColor}{RGB}{33,102,172}
\definecolor{fillColor}{RGB}{33,102,172}

\path[draw=drawColor,line width= 0.4pt,line join=round,line cap=round,fill=fillColor] (258.34,168.36) circle (  3.65);

\path[draw=drawColor,line width= 0.4pt,line join=round,line cap=round,fill=fillColor] (214.44,132.28) circle (  3.65);

\path[draw=drawColor,line width= 0.4pt,line join=round,line cap=round,fill=fillColor] (192.48,131.15) circle (  3.65);

\path[draw=drawColor,line width= 0.4pt,line join=round,line cap=round,fill=fillColor] (170.53, 96.20) circle (  3.65);
\end{scope}
\begin{scope}
\path[clip] (  0.00,  0.00) rectangle (325.21,231.26);
\definecolor{drawColor}{RGB}{0,0,0}

\path[draw=drawColor,line width= 0.4pt,line join=round,line cap=round] ( 53.45, 43.20) -- (309.57, 43.20);

\path[draw=drawColor,line width= 0.4pt,line join=round,line cap=round] ( 53.45, 43.20) -- ( 53.45, 37.80);

\path[draw=drawColor,line width= 0.4pt,line join=round,line cap=round] ( 90.03, 43.20) -- ( 90.03, 37.80);

\path[draw=drawColor,line width= 0.4pt,line join=round,line cap=round] (126.62, 43.20) -- (126.62, 37.80);

\path[draw=drawColor,line width= 0.4pt,line join=round,line cap=round] (163.21, 43.20) -- (163.21, 37.80);

\path[draw=drawColor,line width= 0.4pt,line join=round,line cap=round] (199.80, 43.20) -- (199.80, 37.80);

\path[draw=drawColor,line width= 0.4pt,line join=round,line cap=round] (236.39, 43.20) -- (236.39, 37.80);

\path[draw=drawColor,line width= 0.4pt,line join=round,line cap=round] (272.98, 43.20) -- (272.98, 37.80);

\path[draw=drawColor,line width= 0.4pt,line join=round,line cap=round] (309.57, 43.20) -- (309.57, 37.80);

\node[text=drawColor,anchor=base,inner sep=0pt, outer sep=0pt, scale=  0.90] at ( 53.45, 23.76) {2.0};

\node[text=drawColor,anchor=base,inner sep=0pt, outer sep=0pt, scale=  0.90] at ( 90.03, 23.76) {2.5};

\node[text=drawColor,anchor=base,inner sep=0pt, outer sep=0pt, scale=  0.90] at (126.62, 23.76) {3.0};

\node[text=drawColor,anchor=base,inner sep=0pt, outer sep=0pt, scale=  0.90] at (163.21, 23.76) {3.5};

\node[text=drawColor,anchor=base,inner sep=0pt, outer sep=0pt, scale=  0.90] at (199.80, 23.76) {4.0};

\node[text=drawColor,anchor=base,inner sep=0pt, outer sep=0pt, scale=  0.90] at (236.39, 23.76) {4.5};

\node[text=drawColor,anchor=base,inner sep=0pt, outer sep=0pt, scale=  0.90] at (272.98, 23.76) {5.0};

\node[text=drawColor,anchor=base,inner sep=0pt, outer sep=0pt, scale=  0.90] at (309.57, 23.76) {5.5};

\path[draw=drawColor,line width= 0.4pt,line join=round,line cap=round] ( 43.20, 49.97) -- ( 43.20,219.10);

\path[draw=drawColor,line width= 0.4pt,line join=round,line cap=round] ( 43.20, 49.97) -- ( 37.80, 49.97);

\path[draw=drawColor,line width= 0.4pt,line join=round,line cap=round] ( 43.20,106.34) -- ( 37.80,106.34);

\path[draw=drawColor,line width= 0.4pt,line join=round,line cap=round] ( 43.20,162.72) -- ( 37.80,162.72);

\path[draw=drawColor,line width= 0.4pt,line join=round,line cap=round] ( 43.20,219.10) -- ( 37.80,219.10);

\node[text=drawColor,rotate= 90.00,anchor=base,inner sep=0pt, outer sep=0pt, scale=  0.90] at ( 30.24, 49.97) {0.75};

\node[text=drawColor,rotate= 90.00,anchor=base,inner sep=0pt, outer sep=0pt, scale=  0.90] at ( 30.24,106.34) {0.80};

\node[text=drawColor,rotate= 90.00,anchor=base,inner sep=0pt, outer sep=0pt, scale=  0.90] at ( 30.24,162.72) {0.85};

\node[text=drawColor,rotate= 90.00,anchor=base,inner sep=0pt, outer sep=0pt, scale=  0.90] at ( 30.24,219.10) {0.90};

\path[draw=drawColor,line width= 0.4pt,line join=round,line cap=round] ( 43.20, 43.20) --
	(319.81, 43.20) --
	(319.81,225.86) --
	( 43.20,225.86) --
	cycle;
\end{scope}
\begin{scope}
\path[clip] (  0.00,  0.00) rectangle (325.21,231.26);
\definecolor{drawColor}{RGB}{0,0,0}

\node[text=drawColor,anchor=base,inner sep=0pt, outer sep=0pt, scale=  0.90] at (181.51,  2.16) {Mean citations per response};

\node[text=drawColor,rotate= 90.00,anchor=base,inner sep=0pt, outer sep=0pt, scale=  0.90] at (  8.64,134.53) {$\overline{\mathrm{CG}}$};
\end{scope}
\begin{scope}
\path[clip] ( 43.20, 43.20) rectangle (319.81,225.86);
\definecolor{drawColor}{RGB}{0,0,0}

\node[text=drawColor,anchor=base west,inner sep=0pt, outer sep=0pt, scale=  0.63] at (130.28,196.10) {LEX Chat (RAG)};

\node[text=drawColor,anchor=base,inner sep=0pt, outer sep=0pt, scale=  0.63] at (258.34,175.52) {Haiku 4.5};

\node[text=drawColor,anchor=base west,inner sep=0pt, outer sep=0pt, scale=  0.63] at (221.76,121.70) {Pixtral Large};

\node[text=drawColor,anchor=base east,inner sep=0pt, outer sep=0pt, scale=  0.63] at (185.17,138.31) {Nova Pro};

\node[text=drawColor,anchor=base,inner sep=0pt, outer sep=0pt, scale=  0.63] at (170.53, 83.06) {Nova Lite};
\definecolor{drawColor}{gray}{0.50}

\path[draw=drawColor,line width= 0.4pt,dash pattern=on 4pt off 4pt ,line join=round,line cap=round] (126.62, 69.78) --
	(133.94, 75.31) --
	(141.26, 80.84) --
	(148.58, 86.37) --
	(155.90, 91.90) --
	(163.21, 97.43) --
	(170.53,102.96) --
	(177.85,108.49) --
	(185.17,114.02) --
	(192.48,119.55) --
	(199.80,125.08) --
	(207.12,130.61) --
	(214.44,136.14) --
	(221.76,141.67) --
	(229.07,147.20) --
	(236.39,152.73) --
	(243.71,158.26) --
	(251.03,163.79) --
	(258.34,169.33) --
	(265.66,174.86) --
	(272.98,180.39) --
	(280.30,185.92) --
	(287.62,191.45) --
	(294.93,196.98) --
	(302.25,202.51) --
	(309.57,208.04);
\definecolor{drawColor}{RGB}{33,102,172}
\definecolor{fillColor}{RGB}{33,102,172}

\path[draw=drawColor,line width= 0.4pt,line join=round,line cap=round,fill=fillColor] ( 49.68, 60.48) circle (  1.62);
\definecolor{fillColor}{RGB}{214,39,40}

\path[fill=fillColor] ( 49.68, 54.36) --
	( 51.86, 50.58) --
	( 47.50, 50.58) --
	cycle;
\definecolor{drawColor}{RGB}{0,0,0}

\node[text=drawColor,anchor=base west,inner sep=0pt, outer sep=0pt, scale=  0.72] at ( 56.16, 58.00) {Bedrock (raw)};

\node[text=drawColor,anchor=base west,inner sep=0pt, outer sep=0pt, scale=  0.72] at ( 56.16, 49.36) {RAG-augmented};
\end{scope}
\end{tikzpicture}
\caption{Citation density vs.\ grounding accuracy under $\mathcal{G}_{\text{sparse}}$. The dashed fit is over the four raw Bedrock models only: $\rho = +0.96$ (Spearman $+1.00$). Including the RAG system (triangle) as a fifth point gives Pearson $-0.12$, Spearman $0.00$. With $n = 5$ the sign is not identified; version~1 of this paper reported the five-system Pearson value as though it described the four-model subset.}
\label{fig:density-accuracy}
\end{figure}
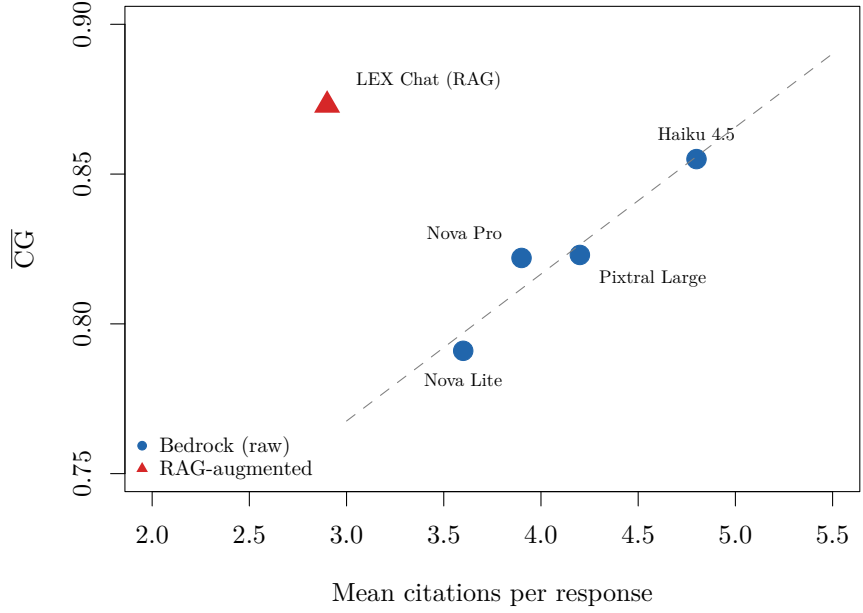

\paragraph{Density--accuracy relationship.}
Version~1 of this paper reported $\rho = -0.12$ for the density--accuracy relationship among raw models.
That value is the Pearson correlation over all \emph{five} systems including the RAG outlier; on the four independent models it is named for, it is $\rho = +0.96$ (Spearman $+1.00$) -- a strong positive relationship of the opposite sign, and the upward-sloping fit in Figure~\ref{fig:density-accuracy} was drawn on that subset all along.
Over all five, Spearman is exactly $0.00$, so the negative Pearson value rests entirely on one leverage point, and that point is the system \S\ref{sec:contamination} shows should never have been in the comparison.
We report the error because it is instructive rather than incidental: at $n = 5$ a single non-independent observation reverses the sign of a correlation, and at $n = 4$ no sign is estimable anyway.
Under $\mathcal{G}_{\text{dense}}$ the question dissolves -- $\CG$ has almost no variance left to correlate with anything.

\subsection{Cross-Domain Analysis}

\begin{table}[t]
\centering
\small
\caption{$\CG$ by legal domain, sparse oracle (mean $\pm$ std) and dense oracle (mean). Under $\mathcal{G}_{\text{dense}}$ every cell is $\geq .96$ and 31 of 35 are exactly $1.00$: the cross-domain structure the sparse oracle appeared to reveal does not survive.}
\label{tab:cg-domains}
\begin{tabular}{@{}lccccc@{}}
\toprule
\textbf{Domain} & \textbf{LEX (RAG)} & \textbf{Haiku~4.5} & \textbf{Pixtral L.} & \textbf{Nova Pro} & \textbf{Nova Lite} \\
\midrule
\multicolumn{6}{@{}l}{\emph{$\mathcal{G}_{\text{sparse}}$}} \\
Constitutional & $1.00 \pm .00$ & $1.00 \pm .00$ & $1.00 \pm .00$ & $1.00 \pm .00$ & $1.00 \pm .00$ \\
Criminal       & $.84 \pm .30$ & $.98 \pm .08$ & $.94 \pm .13$ & $.98 \pm .07$ & $.92 \pm .15$ \\
Civil          & $.99 \pm .03$ & $.92 \pm .20$ & $.88 \pm .28$ & $.93 \pm .15$ & $.75 \pm .36$ \\
Admin.         & $.75 \pm .40$ & $.69 \pm .38$ & $.62 \pm .36$ & $.78 \pm .26$ & $.79 \pm .31$ \\
Military       & $.63 \pm .37$ & $.77 \pm .27$ & $.82 \pm .15$ & $.78 \pm .27$ & $.79 \pm .36$ \\
Family         & $.78 \pm .22$ & $.76 \pm .22$ & $.71 \pm .36$ & $.46 \pm .26$ & $.61 \pm .31$ \\
Labor          & $.90 \pm .13$ & $.71 \pm .26$ & $.64 \pm .25$ & $.49 \pm .34$ & $.62 \pm .27$ \\
\midrule
\multicolumn{6}{@{}l}{\emph{$\mathcal{G}_{\text{dense}}$}} \\
Constitutional & $1.00$ & $1.00$ & $1.00$ & $1.00$ & $1.00$ \\
Criminal       & $1.00$ & $1.00$ & $1.00$ & $1.00$ & $1.00$ \\
Civil          & $1.00$ & $.99$ & $1.00$ & $1.00$ & $.96$ \\
Admin.         & $1.00$ & $1.00$ & $1.00$ & $1.00$ & $1.00$ \\
Military       & $1.00$ & $1.00$ & $1.00$ & $1.00$ & $1.00$ \\
Family         & $1.00$ & $1.00$ & $1.00$ & $1.00$ & $1.00$ \\
Labor          & $1.00$ & $1.00$ & $.98$ & $.99$ & $1.00$ \\
\bottomrule
\end{tabular}
\end{table}

Table~\ref{tab:cg-domains} makes the point sharper than the aggregate does, because the cross-domain pattern was the most substantive-looking finding of the sparse condition.
Under $\mathcal{G}_{\text{sparse}}$ the domains order themselves in a way that invites legal interpretation: constitutional law perfect, criminal law high, family and labor law worst and most variable, military law depressed by wartime statutes not yet in judicial practice.
Under $\mathcal{G}_{\text{dense}}$ that entire structure disappears.
Every cell is at least $.96$; 31 of 35 are exactly $1.00$; military law -- the domain the sparse condition singled out as a genuine coverage gap -- is $1.00$ for every system.

The ordering was a map of harvesting coverage.
Family and labor law scored worst because $\mathcal{G}_{\text{sparse}}$ held few Family Code and Labour Code citations, not because models cite those codes less reliably.
Constitutional law scored perfectly for a different reason again: $\mathcal{G}_{\text{sparse}}$ had no constitutional citation type at all, so the reference implementation admitted every constitutional citation unconditionally, and the ``perfect grounding'' reported in version~1 was a property of the code rather than a measurement.
Against $\mathcal{G}_{\text{dense}}$, where 356 constitutional nodes permit a real lookup, the value is still $1.00$ -- the conclusion happened to be right, but the sparse condition provided no evidence for it.

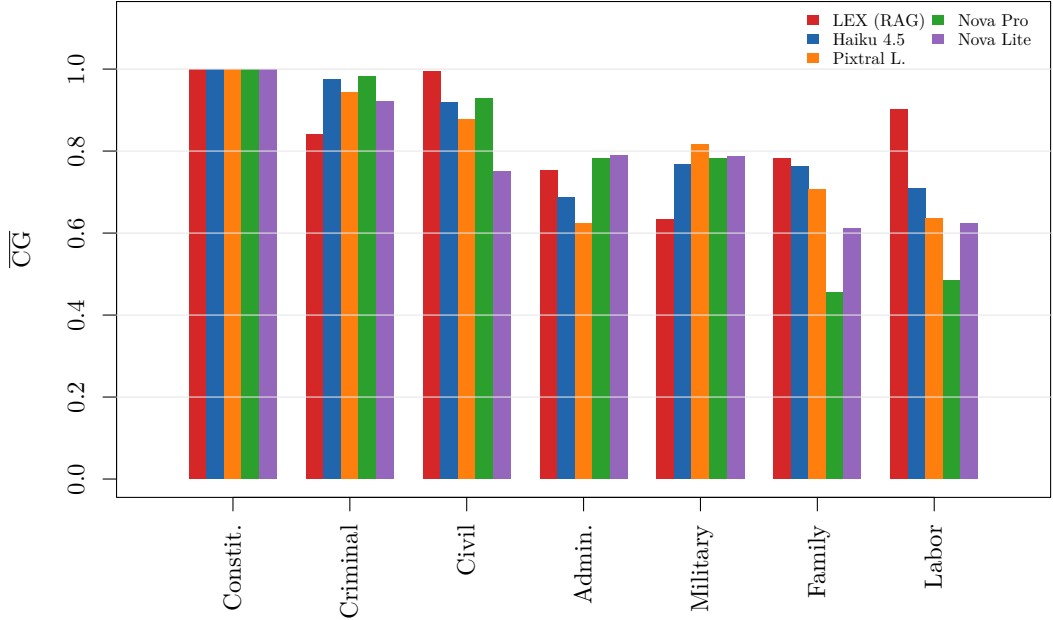
\begin{figure}[t]
\centering
\begin{tikzpicture}[x=1pt,y=1pt]
\definecolor{fillColor}{RGB}{255,255,255}
\path[use as bounding box,fill=fillColor,fill opacity=0.00] (0,0) rectangle (397.48,252.94);
\begin{scope}
\path[clip] (  0.00,  0.00) rectangle (397.48,252.94);
\definecolor{drawColor}{RGB}{0,0,0}

\path[draw=drawColor,line width= 0.4pt,line join=round,line cap=round] ( 40.80, 68.11) -- ( 40.80,222.42);

\path[draw=drawColor,line width= 0.4pt,line join=round,line cap=round] ( 40.80, 68.11) -- ( 35.70, 68.11);

\path[draw=drawColor,line width= 0.4pt,line join=round,line cap=round] ( 40.80, 98.97) -- ( 35.70, 98.97);

\path[draw=drawColor,line width= 0.4pt,line join=round,line cap=round] ( 40.80,129.83) -- ( 35.70,129.83);

\path[draw=drawColor,line width= 0.4pt,line join=round,line cap=round] ( 40.80,160.69) -- ( 35.70,160.69);

\path[draw=drawColor,line width= 0.4pt,line join=round,line cap=round] ( 40.80,191.56) -- ( 35.70,191.56);

\path[draw=drawColor,line width= 0.4pt,line join=round,line cap=round] ( 40.80,222.42) -- ( 35.70,222.42);

\node[text=drawColor,rotate= 90.00,anchor=base,inner sep=0pt, outer sep=0pt, scale=  0.85] at ( 28.56, 68.11) {0.0};

\node[text=drawColor,rotate= 90.00,anchor=base,inner sep=0pt, outer sep=0pt, scale=  0.85] at ( 28.56, 98.97) {0.2};

\node[text=drawColor,rotate= 90.00,anchor=base,inner sep=0pt, outer sep=0pt, scale=  0.85] at ( 28.56,129.83) {0.4};

\node[text=drawColor,rotate= 90.00,anchor=base,inner sep=0pt, outer sep=0pt, scale=  0.85] at ( 28.56,160.69) {0.6};

\node[text=drawColor,rotate= 90.00,anchor=base,inner sep=0pt, outer sep=0pt, scale=  0.85] at ( 28.56,191.56) {0.8};

\node[text=drawColor,rotate= 90.00,anchor=base,inner sep=0pt, outer sep=0pt, scale=  0.85] at ( 28.56,222.42) {1.0};

\path[draw=drawColor,line width= 0.4pt,line join=round,line cap=round] ( 40.80, 61.20) --
	(392.38, 61.20) --
	(392.38,247.84) --
	( 40.80,247.84) --
	cycle;
\end{scope}
\begin{scope}
\path[clip] (  0.00,  0.00) rectangle (397.48,252.94);
\definecolor{drawColor}{RGB}{0,0,0}

\node[text=drawColor,rotate= 90.00,anchor=base,inner sep=0pt, outer sep=0pt, scale=  0.85] at (  8.16,154.52) {$\overline{\mathrm{CG}}$};
\end{scope}
\begin{scope}
\path[clip] ( 40.80, 61.20) rectangle (392.38,247.85);
\definecolor{fillColor}{RGB}{214,39,40}

\path[fill=fillColor] ( 68.12, 68.11) rectangle ( 74.72,222.42);

\path[fill=fillColor] (112.11, 68.11) rectangle (118.71,197.73);

\path[fill=fillColor] (156.10, 68.11) rectangle (162.70,221.34);

\path[fill=fillColor] (200.10, 68.11) rectangle (206.69,184.30);

\path[fill=fillColor] (244.09, 68.11) rectangle (250.69,165.79);

\path[fill=fillColor] (288.08, 68.11) rectangle (294.68,188.93);

\path[fill=fillColor] (332.07, 68.11) rectangle (338.67,207.29);
\definecolor{fillColor}{RGB}{33,102,172}

\path[fill=fillColor] ( 74.72, 68.11) rectangle ( 81.32,222.42);

\path[fill=fillColor] (118.71, 68.11) rectangle (125.31,218.56);

\path[fill=fillColor] (162.70, 68.11) rectangle (169.30,209.92);

\path[fill=fillColor] (206.69, 68.11) rectangle (213.29,174.12);

\path[fill=fillColor] (250.69, 68.11) rectangle (257.29,186.62);

\path[fill=fillColor] (294.68, 68.11) rectangle (301.28,185.85);

\path[fill=fillColor] (338.67, 68.11) rectangle (345.27,177.36);
\definecolor{fillColor}{RGB}{255,127,14}

\path[fill=fillColor] ( 81.32, 68.11) rectangle ( 87.92,222.42);

\path[fill=fillColor] (125.31, 68.11) rectangle (131.91,213.77);

\path[fill=fillColor] (169.30, 68.11) rectangle (175.90,203.28);

\path[fill=fillColor] (213.29, 68.11) rectangle (219.89,164.24);

\path[fill=fillColor] (257.29, 68.11) rectangle (263.88,194.18);

\path[fill=fillColor] (301.28, 68.11) rectangle (307.88,177.05);

\path[fill=fillColor] (345.27, 68.11) rectangle (351.87,166.10);
\definecolor{fillColor}{RGB}{44,160,44}

\path[fill=fillColor] ( 87.92, 68.11) rectangle ( 94.51,222.42);

\path[fill=fillColor] (131.91, 68.11) rectangle (138.51,219.79);

\path[fill=fillColor] (175.90, 68.11) rectangle (182.50,211.46);

\path[fill=fillColor] (219.89, 68.11) rectangle (226.49,188.78);

\path[fill=fillColor] (263.88, 68.11) rectangle (270.48,188.93);

\path[fill=fillColor] (307.88, 68.11) rectangle (314.47,138.32);

\path[fill=fillColor] (351.87, 68.11) rectangle (358.47,142.95);
\definecolor{fillColor}{RGB}{148,103,189}

\path[fill=fillColor] ( 94.51, 68.11) rectangle (101.11,222.42);

\path[fill=fillColor] (138.51, 68.11) rectangle (145.11,210.38);

\path[fill=fillColor] (182.50, 68.11) rectangle (189.10,183.99);

\path[fill=fillColor] (226.49, 68.11) rectangle (233.09,190.01);

\path[fill=fillColor] (270.48, 68.11) rectangle (277.08,189.39);

\path[fill=fillColor] (314.47, 68.11) rectangle (321.07,162.39);

\path[fill=fillColor] (358.47, 68.11) rectangle (365.07,164.24);
\end{scope}
\begin{scope}
\path[clip] (  0.00,  0.00) rectangle (397.48,252.94);
\definecolor{drawColor}{RGB}{0,0,0}

\path[draw=drawColor,line width= 0.4pt,line join=round,line cap=round] ( 84.62, 61.20) -- (348.57, 61.20);

\path[draw=drawColor,line width= 0.4pt,line join=round,line cap=round] ( 84.62, 61.20) -- ( 84.62, 56.10);

\path[draw=drawColor,line width= 0.4pt,line join=round,line cap=round] (128.61, 61.20) -- (128.61, 56.10);

\path[draw=drawColor,line width= 0.4pt,line join=round,line cap=round] (172.60, 61.20) -- (172.60, 56.10);

\path[draw=drawColor,line width= 0.4pt,line join=round,line cap=round] (216.59, 61.20) -- (216.59, 56.10);

\path[draw=drawColor,line width= 0.4pt,line join=round,line cap=round] (260.58, 61.20) -- (260.58, 56.10);

\path[draw=drawColor,line width= 0.4pt,line join=round,line cap=round] (304.58, 61.20) -- (304.58, 56.10);

\path[draw=drawColor,line width= 0.4pt,line join=round,line cap=round] (348.57, 61.20) -- (348.57, 56.10);

\node[text=drawColor,rotate= 90.00,anchor=base east,inner sep=0pt, outer sep=0pt, scale=  0.85] at ( 87.54, 51.00) {Constit.};

\node[text=drawColor,rotate= 90.00,anchor=base east,inner sep=0pt, outer sep=0pt, scale=  0.85] at (131.54, 51.00) {Criminal};

\node[text=drawColor,rotate= 90.00,anchor=base east,inner sep=0pt, outer sep=0pt, scale=  0.85] at (175.53, 51.00) {Civil};

\node[text=drawColor,rotate= 90.00,anchor=base east,inner sep=0pt, outer sep=0pt, scale=  0.85] at (219.52, 51.00) {Admin.};

\node[text=drawColor,rotate= 90.00,anchor=base east,inner sep=0pt, outer sep=0pt, scale=  0.85] at (263.51, 51.00) {Military};

\node[text=drawColor,rotate= 90.00,anchor=base east,inner sep=0pt, outer sep=0pt, scale=  0.85] at (307.50, 51.00) {Family};

\node[text=drawColor,rotate= 90.00,anchor=base east,inner sep=0pt, outer sep=0pt, scale=  0.85] at (351.50, 51.00) {Labor};
\end{scope}
\begin{scope}
\path[clip] ( 40.80, 61.20) rectangle (392.38,247.85);
\definecolor{drawColor}{gray}{0.90}

\path[draw=drawColor,line width= 0.4pt,line join=round,line cap=round] ( 40.80, 98.97) -- (392.38, 98.97);

\path[draw=drawColor,line width= 0.4pt,line join=round,line cap=round] ( 40.80,129.83) -- (392.38,129.83);

\path[draw=drawColor,line width= 0.4pt,line join=round,line cap=round] ( 40.80,160.69) -- (392.38,160.69);

\path[draw=drawColor,line width= 0.4pt,line join=round,line cap=round] ( 40.80,191.56) -- (392.38,191.56);

\path[draw=drawColor,line width= 0.4pt,line join=round,line cap=round] ( 40.80,222.42) -- (392.38,222.42);
\definecolor{fillColor}{RGB}{214,39,40}

\path[fill=fillColor] (300.71,242.49) rectangle (305.00,238.92);
\definecolor{fillColor}{RGB}{33,102,172}

\path[fill=fillColor] (300.71,235.35) rectangle (305.00,231.78);
\definecolor{fillColor}{RGB}{255,127,14}

\path[fill=fillColor] (300.71,228.21) rectangle (305.00,224.64);
\definecolor{fillColor}{RGB}{44,160,44}

\path[fill=fillColor] (347.89,242.49) rectangle (352.17,238.92);
\definecolor{fillColor}{RGB}{148,103,189}

\path[fill=fillColor] (347.89,235.35) rectangle (352.17,231.78);
\definecolor{drawColor}{RGB}{0,0,0}

\node[text=drawColor,anchor=base west,inner sep=0pt, outer sep=0pt, scale=  0.59] at (310.35,238.66) {LEX (RAG)};

\node[text=drawColor,anchor=base west,inner sep=0pt, outer sep=0pt, scale=  0.59] at (310.35,231.52) {Haiku 4.5};

\node[text=drawColor,anchor=base west,inner sep=0pt, outer sep=0pt, scale=  0.59] at (310.35,224.38) {Pixtral L.};

\node[text=drawColor,anchor=base west,inner sep=0pt, outer sep=0pt, scale=  0.59] at (357.53,238.66) {Nova Pro};

\node[text=drawColor,anchor=base west,inner sep=0pt, outer sep=0pt, scale=  0.59] at (357.53,231.52) {Nova Lite};
\end{scope}
\end{tikzpicture}
\caption{$\CG$ by legal domain and model under $\mathcal{G}_{\text{sparse}}$. The cross-domain structure shown here is not reproduced under $\mathcal{G}_{\text{dense}}$ (Table~\ref{tab:cg-domains}, lower block), where all 35 cells lie in $[.96, 1.00]$.}
\label{fig:cg-domains}
\end{figure}

\subsection{Qualitative Analysis}

Four responses from the LEX Chat system, drawn from the released evaluation records, show what the sparse oracle was rejecting.
All four score $\CG = 1.0$ against $\mathcal{G}_{\text{dense}}$.

\paragraph{Contiguous statutory blocks (admin., $\CG_{\text{sparse}} = 0.0$, 6 citations).}
Query: ``\ukr{Як регулюється відповідальність за порушення митних правил?}'' (Liability for customs violations).
The model cites Art.~461--466 of the Customs Code -- the six consecutive articles that constitute the chapter on liability for customs offences.
$\mathcal{G}_{\text{sparse}}$ contained none of them, so all six were recorded as hallucinations.

\paragraph{Wartime criminal law (criminal, $\CG_{\text{sparse}} = 0.25$, 12 citations).}
Query: ``\ukr{Які підстави для звільнення від покарання у зв'язку із закінченням строків давності?}'' (Grounds for release from punishment on limitation).
Nine of twelve citations were flagged, among them Art.~109, 110, 112, 113 (actions against the constitutional order), 437, 438, 439 and 442 (war crimes and genocide) of the Criminal Code -- the provisions under which Ukrainian courts have been prosecuting since 2022, and precisely the articles for which the limitation question is asked.
These are among the most heavily litigated provisions in the corpus; their absence from $\mathcal{G}_{\text{sparse}}$ reflects what had been harvested, nothing about the model.

\paragraph{Procedural review (admin., $\CG_{\text{sparse}} = 0.0$, 3 citations).}
Query: ``\ukr{Які підстави для перегляду судового рішення за нововиявленими обставинами?}'' (Grounds for review on newly discovered circumstances).
The model cites Art.~361--363 of the Code of Administrative Procedure -- the three articles that define exactly this procedure.

\paragraph{A single general provision (criminal, $\CG_{\text{sparse}} = 0.0$, 1 citation).}
Query: ``\ukr{Які права підозрюваного при затриманні?}'' (Rights of a suspect upon detention).
The response yields one extractable citation, Art.~42 of the Criminal Procedure Code -- the article that enumerates the rights of a suspect.
Flagged by $\mathcal{G}_{\text{sparse}}$, the response scored $\CG = 0.0$: a single lookup miss drove a response-level score to zero, which is what the per-response averaging in Definition~\ref{def:cg} does whenever density is low.

\medskip
\noindent
The pattern across these cases is not random noise but systematic omission of whole statutory neighbourhoods.
A metric that rejects Art.~461--466 of the Customs Code and Art.~437--442 of the Criminal Code is not detecting fabrication; it is reporting which parts of the law had been indexed.

\subsection{Coverage Alone Reproduces the Sparse Snapshot}
\label{sec:sweep}

The two snapshots differ in coverage and in type inventory (\S\ref{sec:snapshots}), so on their own they cannot tell us which caused the shift.
Subsampling settles it: within $\mathcal{G}_{\text{dense}}$ the type inventory is fixed by construction, and only the harvested fraction varies.

\paragraph{Method.}
We have per-node record counts but not the record list, so we model harvesting as uniform independent sampling of citation records at rate $p$.
Write $k(v)$ for the number of records naming statute node $v$ in $\mathcal{G}_{\text{dense}}$.
Node $v$ is present in the sampled oracle iff at least one of those records survives:
\begin{equation}
  \Pr\bigl[\,v \in V_n^{(p)}\,\bigr] = 1 - (1-p)^{k(v)} .
  \label{eq:survival}
\end{equation}
Rare nodes therefore vanish first, which is the mechanism we claim drives the effect.
We sweep $p$ over 18 rates spanning $10^{-4}$ to $1$, five seeds each, and re-score the same responses with the same extractor and metric.
The model is calibrated on nothing but the record count.

\begin{figure}[t]
\centering
\input{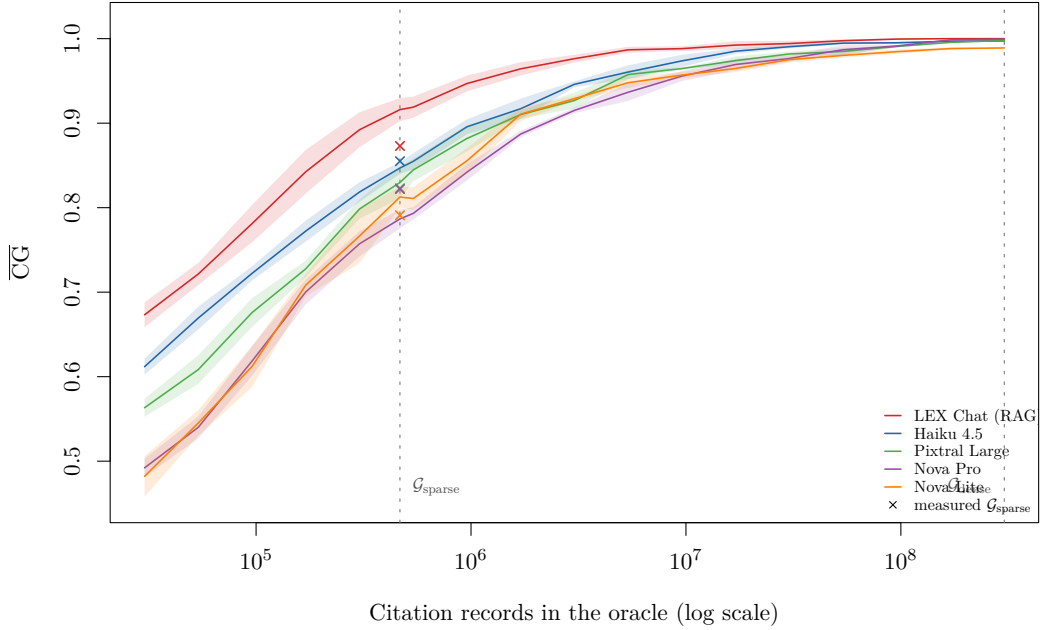}
\caption{$\CG$ against oracle size under Equation~\ref{eq:survival}, five seeds per rate, bands showing $\pm 1$ s.d. Crosses mark the values actually measured against $\mathcal{G}_{\text{sparse}}$, which the sweep was not fitted to.}
\label{fig:coverage-sweep}
\end{figure}

\paragraph{It reconstructs the lost snapshot.}
Evaluated at the sparse snapshot's record count, the sweep predicts $\CG$ within a mean absolute error of $0.018$ (Figure~\ref{fig:coverage-sweep}, crosses), and places two of the four systems in their measured rank positions -- Haiku first and Pixtral second, with Nova Pro and Nova Lite transposed.
A model that knows only how often each statute is cited, and nothing whatever about the language models, therefore accounts for the level of the sparse condition's scores.
It does not need to account for their order, because \S\ref{sec:separability} shows there was no order to account for.

\paragraph{How dense is dense enough.}
The sweep converts the qualitative worry into a threshold.
For all four systems to fall within $0.05$ of their full-coverage $\CG$ takes ${\approx}9.6 \times 10^6$ records ($3.2\%$ of the graph); within $0.02$ takes ${\approx}5.4 \times 10^7$ ($17.8\%$); within $0.01$, ${\approx}9.6 \times 10^7$ ($31.6\%$).
$\mathcal{G}_{\text{sparse}}$, at $4.7 \times 10^5$ records, sits a factor of twenty below even the loosest of these, with a gap of $0.21$ to the asymptote.

\paragraph{What the model assumes.}
Real harvesting is neither independent per record nor uniform over the corpus: a decision contributes several citations at once, and crawls are ordered by court, by year, or by whatever was reachable first.
Equation~\ref{eq:survival} ignores both, which will understate the variance at a given coverage and may misplace the curve for non-uniform harvest orders.
A by-court or by-year subsampling scheme is the robustness check we have not run.
The reconstruction in Figure~\ref{fig:coverage-sweep} is evidence that the approximation is not badly wrong at the one point where we can check it.

\subsection{No Ranking Was Ever Resolvable}
\label{sec:separability}

Everything so far concerns the \emph{level} of $\CG$, and shows it is set by the oracle.
A separate question is whether $\CG$ ever supported the other thing version~1 reported: an ordering of systems.
That question needs an uncertainty estimate over the query sample, which no version of this work has provided.

We bootstrap mean $\CG$ over the 100 queries ($B = 5000$ resamples) and ask, for every pair of systems, whether the difference clears a 95\% interval.

\begin{table}[t]
\centering
\small
\caption{Bootstrap over the 100 evaluation queries ($B = 5000$). Separable pairs = pairs of systems whose $\CG$ difference clears a 95\% interval, out of the $\binom{4}{2} = 6$ pairs. No pair separates at any oracle size.}
\label{tab:separability}
\begin{tabular}{@{}rrrr@{}}
\toprule
\textbf{Records in oracle} & $\overline{\CG}$ & \textbf{Spread} & \textbf{Separable pairs} \\
\midrule
$4.7 \times 10^5$ (${=}\,\mathcal{G}_{\text{sparse}}$) & 0.826 & 0.035 & 0 / 6 \\
$3.0 \times 10^6$   & 0.928 & 0.051 & 0 / 6 \\
$9.6 \times 10^6$   & 0.964 & 0.018 & 0 / 6 \\
$3.0 \times 10^7$   & 0.979 & 0.011 & 0 / 6 \\
$9.6 \times 10^7$   & 0.993 & 0.016 & 0 / 6 \\
$3.0 \times 10^8$ (${=}\,\mathcal{G}_{\text{dense}}$) & 0.996 & 0.010 & 0 / 6 \\
\bottomrule
\end{tabular}
\end{table}

Table~\ref{tab:separability} gives the answer: \textbf{zero}, at every coverage from the sparse snapshot to the full graph.
Against $\mathcal{G}_{\text{dense}}$ every system's 95\% interval reaches 1.0000 -- Nova Pro $[0.9979, 1.0000]$, Pixtral $[0.9932, 1.0000]$, Haiku $[0.9918, 1.0000]$, Nova Lite $[0.9655, 1.0000]$ -- and pairwise $\Pr[A > B]$ never leaves the range $0.24$ to $0.63$.

This reframes the ranking result of \S\ref{sec:evaluation}.
We reported there that the order inverts between the two oracles.
It does, but neither order was resolvable, so the inversion is not evidence about the systems; it is two draws from the same absence of signal.
The measured sparse condition already contained the tell: Pixtral and Nova Pro, adjacent in that ranking, differ by $0.001$.

\paragraph{The dilemma.}
Put the two halves together and the metric is caught between them.
Where the oracle is sparse, $\CG$ varies across systems -- but \S\ref{sec:sweep} shows the variation tracks citation frequency rather than model quality, and \S\ref{sec:validation} shows the flags it produces are false positives at a rate of $54/54$.
Where the oracle is dense enough for its flags to mean something, the spread collapses to $0.010$ and nothing separates.
There is no coverage at which existence-based citation grounding is both valid and discriminative.
Densifying the graph, the obvious remedy for the first horn, is what produces the second.

\paragraph{What this does not show.}
Separability depends on sample size as well as on the metric, and 100 queries is a small sample.
A larger query set would eventually resolve differences of $0.008$ if they are real.
The claim is therefore about the evaluation as it is normally practised -- at the query counts these studies actually use, this metric does not rank systems -- and not a proof that no difference exists.
It is also specific to mean $\CG$ over responses; a metric aggregating over citations rather than responses would have different variance, though the saturation at high coverage would remain.

\subsection{A System That Shares a Corpus With the Oracle}
\label{sec:contamination}

LEX Chat retrieves over the EDRSR corpus and is then scored against a graph extracted from that same corpus.
The metric rewards it for overlap between its retrieval index and the ground truth, which is not the property $\CG$ is supposed to measure.

The size of the artefact is visible in Table~\ref{tab:cg-results}.
Under $\mathcal{G}_{\text{sparse}}$ it leads every independent system, at the lowest citation density (2.9); version~1 read this as evidence that retrieval augmentation grounds citations in real judicial practice and recommended it over model scaling.
Under $\mathcal{G}_{\text{dense}}$ its margin over the weakest system is $0.011$, and what distinguishes it is not accuracy but abstention: it declines to cite on 21 of 100 queries against 6--9 for the others.

Retrieval did constrain this system to cite statutes with harvested judicial support.
That is exactly the property a sparse oracle rewards, and exactly why the reward is circular.
An evaluation in which the graph is both the ground truth and an input to one of the evaluated systems cannot separate grounding from overlap, and we can see no way to repair the comparison short of an oracle built from a disjoint source.

Excluding it costs little and buys clarity: the headline moves from $13$--$21\%$ to $15$--$21\%$, the coverage and separability results are untouched, and the density--accuracy correlation stops depending on which subset one happens to fit.

\subsection{Which Oracle Is Right?}
\label{sec:validation}

Disagreement between two oracles does not by itself tell us which one is closer to the truth.
We adjudicate against an independent source: the \texttt{legislation\_articles} registry, which indexes statute texts directly rather than by observed judicial citation, and is therefore not derived from the graph under test.

\paragraph{The sparse oracle's flags.}
Version~1 of this paper cross-referenced all 54 unique citations that $\mathcal{G}_{\text{sparse}}$ flagged for the RAG system and found that \emph{all 54 name real, existing statute articles}.
No fabricated provision was among them.
That finding was reported in version~1 and is reproduced here unchanged; what changed is its interpretation.
Version~1 read a 100\% false-positive rate as evidence that $\CG$ is a \emph{conservative} detector -- one that over-flags but never misses -- and concluded that the asymmetry was desirable for legal use.
That inference does not hold.
A 100\% false-positive rate is not evidence of a favourable trade-off; it is evidence that the flags carried no information about fabrication at all.
Nothing in the sparse experiment constrained the false-negative rate in either direction, so ``no false negatives'' was an assumption presented as a result.

\paragraph{The dense oracle's flags.}
$\mathcal{G}_{\text{dense}}$ rejects four of the 1{,}108 distinct pairs.
Cross-referenced against the same registry:

\begin{itemize}[leftmargin=*, nosep]
  \item \textbf{Art.~393 of the Housing Code} -- the Housing Code's last article is 193, and the registry indexes 189 numeric articles with no gaps in the closing range. A fabrication.
  \item \textbf{Art.~1522 of the Civil Code} -- the Civil Code's last article is 1308. A fabrication.
  \item \textbf{Art.~210 of the Labour Code} -- present and current in the registry. A coverage gap: a real article that no harvested decision cites.
  \item \textbf{Art.~261 of the Labour Code} -- we cannot classify it. See below.
\end{itemize}

\paragraph{The case we could not resolve.}
Labour Code Art.~261 sits \emph{inside} its code's range: the last article is 265, and 263 and 265 are both indexed.
It appears in none of the registry's 37{,}018 rows for that code -- neither the 249 current ones nor the 36{,}769 historical ones -- and its neighbours 262 and 264 are missing in the same way, while the Housing Code shows no such gaps anywhere.
A cluster of holes below the last article, in a code whose surroundings are dense, is what a repealed provision looks like: articles excluded from the consolidated text carry no body in any edition, so a parser that requires text drops them from all of them.
That reading is consistent with the evidence and inconsistent with the model having invented the number.
But it cannot be established from this registry, because absence from every edition is equally consistent with incomplete harvesting of the historical text.

We report it as indeterminate, and note that this is the paper's argument arriving in miniature.
If Art.~261 was repealed, a model citing it committed the second failure mode listed in \S\ref{sec:intro} -- an outdated norm, not a fabricated one -- and that is a $\CT$ failure, the component nothing in this line of work computes.
Existence checking cannot separate fabrication from repeal from missing index.
Neither, on this evidence, could we.

\noindent
Confirmed fabrications are therefore \emph{two} of four dense-oracle flags, against zero of 54 for the sparse oracle, with one indeterminate and one coverage gap.
Flag precision rises from $0/54$ to at least $2/4$ as the graph densifies, while the number of flags falls by an order of magnitude.

\paragraph{What this establishes.}
The dense reading is the correct one, and existence checking is a real -- if narrow -- capability: with a sufficiently covered graph it isolates fabricated article numbers at high precision.
But the quantity version~1 reported was not that capability.
``13--21\% of citations hallucinated'' was, on this evidence, at most a fraction of a percent of hallucination and the rest missing index.
The two are indistinguishable from inside the metric, which is why the sparse result was reportable in the first place.

\section{Citation Grounding DPO}
\label{sec:cg-dpo}

The preceding sections establish what $\CG$ can and cannot measure.
A separate question is whether the graph, whatever its value as an evaluation instrument, carries enough signal to serve as a \emph{training} oracle -- constructing preference pairs without annotators.
We report such an experiment and, having just shown how easily a graph-derived number can be over-read, are correspondingly careful about what its result licenses.

\subsection{Motivation}

Standard RLHF~\citep{ouyang2022training} and DPO~\citep{rafailov2023direct} rely on human-generated preference pairs.
In the legal domain, this requires annotators with domain expertise -- practicing lawyers who can distinguish correct from hallucinated citations.
Such annotation is expensive, does not scale, and is unavailable for low-resource legal systems.

The citation graph provides an alternative: for any court decision $d \in V_d(\mathcal{G})$ with citation vector $\mathbf{c}(d) = [c_1, \ldots, c_m]$, we can algorithmically construct corrupted versions $\tilde{\mathbf{c}}$ that violate specific $\CG$ components.
The original text with correct citations becomes the preferred response; the corrupted version becomes the rejected response.

\subsection{Corruption Strategies}

\begin{definition}[CG-DPO Pair Construction]
\label{def:cg-dpo}
Let $d$ be a court decision with citation vector $\mathbf{c}(d) = [c_1, \ldots, c_m]$.
Define a corruption function $\sigma : \mathbf{c} \to \tilde{\mathbf{c}}$ that systematically modifies citations.
A CG-DPO pair is:
\begin{equation}
  (x,\; y_w,\; y_l) = \bigl(\text{prompt}(d),\; \text{text}(d, \mathbf{c}),\; \text{text}(d, \sigma(\mathbf{c}))\bigr),
  \label{eq:cg-dpo-pair}
\end{equation}
where $y_w$ is the original text with correct citations (preferred) and $y_l$ is the text with corrupted citations (rejected).
\end{definition}

The implementation applies one of four named strategies, chosen uniformly at random per pair.
We describe them as implemented, which differs in one respect from how they were described in version~1:

\begin{enumerate}[leftmargin=*, nosep]
  \item \textbf{\texttt{article\_swap}} -- replace an article number with a different article of the same codex (e.g., Art.~625 CC $\to$ Art.~130 CC). Targets $\CP$: the cited article exists but is wrong.

  \item \textbf{\texttt{law\_swap}} -- replace the codex, preserving the article number (e.g., Art.~625 CC $\to$ Art.~625 Criminal Code). Targets $\CR$: the article exists but in an unrelated legal domain.

  \item \textbf{\texttt{hallucination}} -- replace a real article with a number drawn from $[9000, 9999]$ and checked to be absent from the codex. Targets $\CP$.

  \item \textbf{\texttt{anachronism}} -- \emph{as implemented}, replaces the citation with a random article of a random other codex. This is operationally identical to \texttt{law\_swap} and involves no dates. Version~1 described it as replacing a statute with one enacted after the decision date, targeting $\CT$; no such logic exists in the code, and $\CT$ is not exercised by any strategy.
\end{enumerate}

The dataset therefore contains three distinct corruption types, not four, and covers two of the three $\CG$ components.
We keep the four-way naming here so that the released dataset's \texttt{strategy} field remains interpretable.

\subsection{Dataset Construction}

Pairs were drawn from decisions of a single year, 2024, that contain at least 3 distinct legislation citations each, yielding 2{,}244 decisions and one pair per decision per Definition~\ref{def:cg-dpo}.
The single-year restriction is an implementation choice, not a sampling design; it further limits any claim about temporal generality.
The dataset was split 80/10/10: 1{,}795 training pairs, 224 validation pairs, 225 test pairs.

Both $y_w$ and $y_l$ are generated from fixed Ukrainian templates that differ only in the corrupted citation, so the pair is separable by surface form alone.
We return to this in \S\ref{sec:dpo-caveat}.

\subsection{Training}

\paragraph{Model.} Qwen2.5-3B-Instruct~\citep{yang2024qwen25} with LoRA adaptation~\citep{hu2022lora}: rank $r = 16$, $\alpha = 32$, maximum sequence length 2{,}048.

\paragraph{Optimization.} DPO loss~\citep{rafailov2023direct}:
\begin{equation}
  \mathcal{L}_{\text{DPO}}(\theta) = -\mathbb{E}_{(x, y_w, y_l) \sim \mathcal{P}} \left[ \log \sigma \left( \beta \log \frac{p_\theta(y_w | x)}{p_{\text{ref}}(y_w | x)} - \beta \log \frac{p_\theta(y_l | x)}{p_{\text{ref}}(y_l | x)} \right) \right],
  \label{eq:dpo-loss}
\end{equation}
where $p_{\text{ref}} = p_{\theta_0}$ is the frozen base model (Qwen2.5-3B-Instruct), $\beta = 0.1$.

\paragraph{Hyperparameters.} 3 epochs, batch size 1, gradient accumulation 16 (effective batch size 16), learning rate $5 \times 10^{-5}$, cosine schedule, 339 optimizer steps.
Trained on an NVIDIA H100 80GB with 3 seed values (42, 123, 456) for stability evaluation.

\paragraph{Correction to version~1.}
Version~1 reported this experiment as Qwen2.5-7B-Instruct at $\beta = 0.01$ and learning rate $2 \times 10^{-5}$, with 98.5\% mean accuracy and rewards margin $+14.9$.
No run with those settings exists.
The configuration and metrics given here are those of the logged run (tracking-server experiment \texttt{cg-dpo-citation-grounding}, runs \texttt{cg-dpo-seed\{42,123,456\}}), and match the released per-seed summary artefact.

\subsection{Results}

\begin{table}[t]
\centering
\small
\caption{CG-DPO training results across 3 seeds (Qwen2.5-3B-Instruct, LoRA). Accuracy = correct classification of preferred vs.\ rejected responses on the 224-pair validation split.}
\label{tab:cg-dpo-results}
\begin{tabular}{@{}lrrrr@{}}
\toprule
\textbf{Metric} & \textbf{Seed 42} & \textbf{Seed 123} & \textbf{Seed 456} & \textbf{Mean} \\
\midrule
Train accuracy     & 100\% & 100\% & 100\% & 100\% \\
Eval loss          & 0.0063 & 0.0031 & 0.0052 & 0.0049 \\
Eval accuracy      & 99.6\% & 100\% & 99.6\% & 99.7\% \\
Rewards margin     & $+29.0$ & $+24.6$ & $+24.2$ & $+25.9$ \\
Training time (s)  & 1{,}852 & 1{,}859 & 1{,}857 & 1{,}856 \\
\bottomrule
\end{tabular}
\end{table}

Table~\ref{tab:cg-dpo-results} presents results.
Validation accuracy is 99.7\% on average (223/224, 224/224, 223/224 across seeds), training accuracy saturates at 100\%, and cross-seed spread is under 0.3~pp.

\begin{figure}[t]
\centering
\input{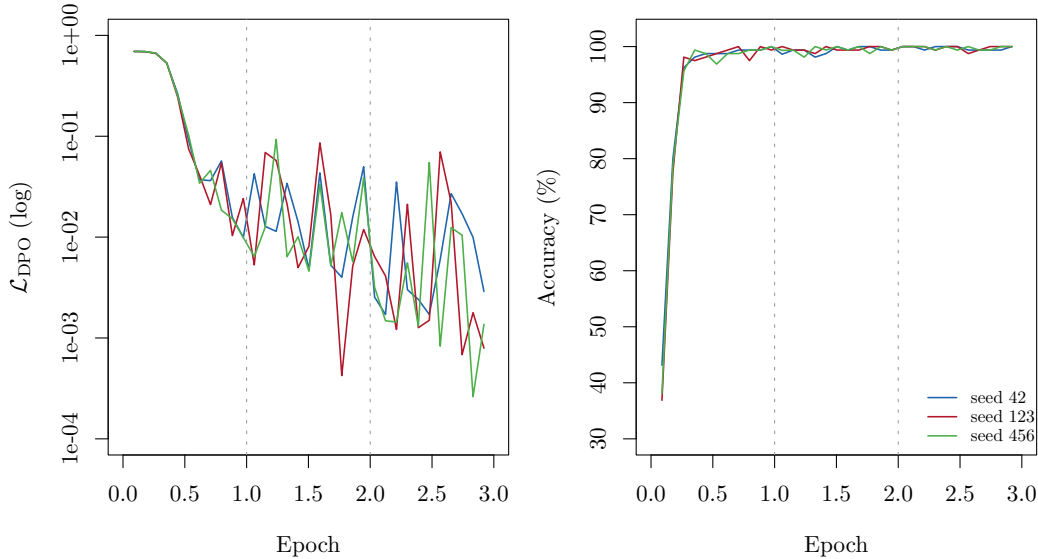}
\caption{Convergence of CG-DPO training (Qwen2.5-3B-Instruct, 3 seeds, 339 logged steps). Left: DPO loss (log scale). Right: classification accuracy of preferred vs.\ rejected pairs. Vertical dashed lines mark epoch boundaries. Accuracy exceeds 95\% by step 30, before a quarter of the first epoch has elapsed -- see \S\ref{sec:dpo-caveat}.}
\label{fig:convergence}
\end{figure}

\paragraph{Convergence dynamics.}
Figure~\ref{fig:convergence} shows the trajectory: near-chance at step 10, above 95\% by step 30 -- roughly a quarter of the first epoch -- and saturated thereafter.

\subsection{What This Result Does Not Show}
\label{sec:dpo-caveat}

The natural reading of Table~\ref{tab:cg-dpo-results} is that graph-derived supervision substitutes for legal expertise.
The evaluation in this paper is a caution against exactly that kind of inference, and it applies here too.

Chosen and rejected responses are generated from the same fixed template and differ in a single citation.
Under \texttt{hallucination} -- one strategy in four, so roughly a quarter of pairs -- the rejected response contains an article number in $[9000, 9999]$, a range in which no Ukrainian codex has provisions.
A model can separate those pairs by numeral length alone.
Under \texttt{law\_swap} and \texttt{anachronism} -- together another half of the pairs, and operationally the same strategy -- the corrupted citation attaches an article number to an unrelated codex, detectable from the mismatch between the named code and the surrounding boilerplate without evaluating whether the provision is apposite.
Saturation within a quarter of an epoch is consistent with the task being solved this way.

The honest statement is therefore narrow: the graph supplies enough signal to train a discriminator on this construction of the task, and 99.7\% measures performance on that construction.
It is not evidence that the model learned which statute governs a dispute, and version~1's claim that the graph is ``an effective algorithmic substitute for human oversight'' is not supported by it.
Establishing that would require corruptions that are hard by design -- adjacent articles within the same chapter, provisions repealed at the decision date -- and an end-to-end evaluation on open generation rather than pairwise discrimination.
We report the experiment because the negative framing matters: the same graph that cannot support the evaluation claim in \S\ref{sec:evaluation} also cannot, on this evidence, support the alignment claim.

\section{Discussion}
\label{sec:discussion}

\paragraph{Why the sparse result was believable.}
It is worth being precise about why an artefact survived to publication, because none of the individual steps were unusual.
The reported range (13--21\%) fell close to an independently published rate for a different jurisdiction and method~\citep{magesh2024hallucination}, which read as convergent validity.
The cross-domain ordering admitted a fluent legal explanation at every position: the Constitution is cited everywhere, wartime statutes are new, family and labour disputes are settled below the cassation threshold.
The RAG system, whose retrieval index shares a corpus with the oracle, came out on top, which matched the prior expectation that retrieval helps.
And the metric had been validated -- against a legislation registry, in a check that returned a 100\% false-positive rate and was read as evidence of conservatism rather than of failure.
Every check that was run returned a plausible answer.
The check that was not run was the one that varies the oracle while holding everything else fixed.

\paragraph{A coverage condition, in measured form.}
$\CG$ is interpretable only where the graph is dense enough that absence is informative.
Equation~\ref{eq:survival} says why: a node cited $k$ times survives harvesting at rate $p$ with probability $1-(1-p)^k$, so at small $p$ the metric reports the citation frequency of a provision rather than its existence.
\S\ref{sec:sweep} turns this into numbers for our corpus -- ${\approx}9.6 \times 10^6$ records for all systems to sit within $0.05$ of their asymptote, ${\approx}5.4 \times 10^7$ for $0.02$ -- and shows $\mathcal{G}_{\text{sparse}}$ sitting twenty times below the looser threshold.
The thresholds themselves are corpus-specific; the shape is not, since it follows from Equation~\ref{eq:survival} and the citation-frequency distribution of any legal corpus, which is heavy-tailed everywhere it has been measured.

The practical corollary for anyone building such a metric: report $|E|$, $|V_d|$ and the harvesting date alongside every score, and treat scores as incomparable across snapshots.
A grounding number without its oracle's coverage is not interpretable, in the same way that a p-value without a sample size is not.

\paragraph{Where discriminative signal must come from.}
Once the graph is dense, existence checking saturates: 1{,}104 of 1{,}108 pairs pass, and no ranking of systems survives.
This is not a defect to be fixed by growing the graph further -- it is the limit behaviour of the definition.
Legal citation errors that matter in practice are mostly not fabrications of article numbers.
They are correct articles cited for the wrong proposition, and provisions that were valid once but not at the material date.
Those are the $\CR$ and $\CT$ components of Definition~\ref{def:components}, which this line of work -- including version~1 of this paper -- has defined but never computed.
Version~1 nonetheless described its examples as demonstrating ``the diagnostic power of the three-component decomposition''; no component other than existence was ever evaluated.
We regard operationalising $\CR$ and $\CT$ as the substantive open problem, and the existence component as essentially solved and essentially uninformative.

\paragraph{Limitations.}

\begin{itemize}[leftmargin=*, nosep]
  \item \textbf{The sweep is a model, not a replay.} \S\ref{sec:sweep} samples citation records independently and uniformly. Real harvesting is clustered by decision and ordered by court, year or crawl reachability. The reconstruction of the sparse condition is evidence the approximation is serviceable at one point; a by-court or by-year scheme is the robustness check we have not run.

  \item \textbf{The sparse snapshot is gone.} $\mathcal{G}_{\text{sparse}}$ was overwritten in place. Its record count is the figure quoted in version~1 and its decision count is a later approximation; neither is a re-measurement, and its side of Table~\ref{tab:cg-results} cannot be independently recomputed. The dense side can, and \S\ref{sec:sweep} approximates the sparse side from surviving data.

  \item \textbf{Extraction covers 16 codexes plus the Constitution.} Citations to named laws are not extracted, so no model is ever charged for fabricating one -- even though named-law articles are the largest node class in $\mathcal{G}_{\text{dense}}$ ($279{,}735$ nodes). Measured $\CG$ is thus conditional on a citation form the extractor recognises, and the dense-side ceiling would likely be lower with broader extraction.

  \item \textbf{Node hygiene in $\mathcal{G}_{\text{dense}}$.} The dense snapshot is machine-extracted and not clean: its \texttt{law\_number} field contains $127{,}175$ distinct values, many of them fragments of decision text, and its 356 constitutional nodes include article numbers outside the Constitution's range of 1--161. This inflates $|V_n|$ and can only make the dense condition \emph{more} permissive, so it does not threaten the direction of our result -- but it does mean the dense oracle over-accepts, and the true fabrication rate lies at or above the 4 pairs we report.

  \item \textbf{Sample size cuts both ways.} 100 queries, four independent systems, one jurisdiction. The separability result in \S\ref{sec:separability} is partly a statement about this sample: a larger query set would eventually resolve differences of $0.008$ if they are real. What we can say is that at the query counts these evaluations actually use, the metric does not rank systems at any oracle coverage. No positive claim in this paper depends on separating systems.

  \item \textbf{Jurisdiction.} All experiments are on Ukrainian law -- a codified system. In common-law systems the primary citation object is case law, so $V_n$ would be prior decisions; we expect oracle dependence to be more severe there, since case-law coverage is patchier than statute coverage, but we have not tested this.

  \item \textbf{CG-DPO scope.} The 99.7\% accuracy measures pairwise discrimination on a construction that is largely solvable from surface form (\S\ref{sec:dpo-caveat}), on decisions from a single year. It supports no claim about open-ended generation.
\end{itemize}

\section{Conclusion}
\label{sec:conclusion}

Verifying generated legal citations against a graph of real judicial practice is an attractive idea, and it works -- but it measures the graph at least as much as it measures the model.
Holding the responses, the extractor and the metric fixed, and changing which snapshot of one national citation graph answers the membership query, moves the reported hallucination rate from 15--21\% to 0.1--1.1\% and does not preserve the ranking of the four evaluated systems.
Subsampling shows the cause is coverage: a model of harvesting calibrated on nothing but the record count reproduces the sparse snapshot's scores to within $0.018$, without knowing anything about the language models at all.
Bootstrapping over the queries shows the ordering was never resolvable -- no pair of systems separates at 95\% at any oracle size we tested.
An independent legislation registry settles which reading is right: all 54 citations flagged by the sparse oracle name real statute articles, while two of the four flagged by the dense oracle name articles beyond their code's last provision.
The sparse oracle was not a conservative detector, as we previously argued; it was a coverage probe misread as one.

The two findings together are worse for the method than either alone, and this is the paper's main claim.
A sparse oracle discriminates between systems, but what it discriminates is how much of the corpus has been harvested.
A dense oracle produces trustworthy verdicts on individual citations, and separates nothing.
Densifying the graph -- the obvious response to the first problem -- is what creates the second.
On this evidence there is no coverage at which existence-based citation grounding is both valid and discriminative.

What survives is narrow and worth stating plainly.
Existence checking against a densely covered graph is a high-precision, low-recall detector of fabricated article numbers, and nothing more.
It cannot rank systems, its output is not a hallucination rate, and a score computed against one snapshot is not comparable to a score computed against another.
Any grounding metric of this family should be reported with its oracle's edge count, decision coverage and harvesting date, or it cannot be interpreted.

The errors that matter in legal practice -- a real article cited for a proposition it does not support, a provision that was in force once but not on the material date -- are the relevance and temporality components that this line of work has repeatedly defined and never computed.
That is where the problem actually is.
We release the dense snapshot's node set, the responses, the extracted citations, the subsampling code and the evaluation code, so that the comparison reported here can be checked and the next such metric can be tested against its oracle before it is trusted.
The sparse snapshot cannot be released: it was overwritten in place ten weeks after the evaluation, by a routine re-derivation that discarded nothing anyone would have thought worth keeping.
The citation graph, evaluation framework, and CG-DPO dataset are released as open resources.\footnote{Citation graph: \url{https://huggingface.co/datasets/overthelex/ua-court-citation-graph}. Code and data: \url{https://huggingface.co/datasets/overthelex/citation-grounding-eval}.}

\bibliographystyle{plainnat}
\bibliography{references}

\begin{thebibliography}{24}
\providecommand{\natexlab}[1]{#1}
\providecommand{\url}[1]{\texttt{#1}}
\expandafter\ifx\csname urlstyle\endcsname\relax
  \providecommand{\doi}[1]{doi: #1}\else
  \providecommand{\doi}{doi: \begingroup \urlstyle{rm}\Url}\fi

\bibitem[Bommarito and Katz(2010)]{bommarito2010mathematical}
Michael~J Bommarito and Daniel~Martin Katz.
\newblock A mathematical approach to the study of the {United States Code}.
\newblock \emph{Physica A}, 389\penalty0 (19):\penalty0 4195--4200, 2010.

\bibitem[Bose(2026)]{bose2026falkor}
Joy Bose.
\newblock {Falkor-IRAC}: Graph-constrained generation for verified legal reasoning in {Indian} judicial {AI}.
\newblock \emph{arXiv preprint arXiv:2605.14665}, 2026.

\bibitem[Chalkidis et~al.(2022)Chalkidis, Jana, Hartung, Bommarito, Androutsopoulos, Katz, and Aletras]{chalkidis2021lexglue}
Ilias Chalkidis, Abhik Jana, Dirk Hartung, Michael Bommarito, Ion Androutsopoulos, Daniel~Martin Katz, and Nikolaos Aletras.
\newblock {LexGLUE}: A benchmark dataset for legal language understanding in {English}.
\newblock In \emph{Proceedings of ACL}, 2022.

\bibitem[Choi et~al.(2023)Choi, Hickman, Monahan, and Schwarcz]{choi2023chatgpt}
Jonathan~H Choi, Kristin~E Hickman, Amy Monahan, and Daniel Schwarcz.
\newblock {ChatGPT} goes to law school.
\newblock \emph{Journal of Legal Education}, 71\penalty0 (3), 2023.

\bibitem[Colombo et~al.(2024)Colombo, Pires, Vieira, et~al.]{colombo2024saullm}
Pierre Colombo, Telmo Pires, Rui Vieira, et~al.
\newblock {SaulLM-7B}: A pioneering large language model for law.
\newblock \emph{arXiv preprint arXiv:2403.03883}, 2024.

\bibitem[Conneau et~al.(2020)Conneau, Khandelwal, Goyal, Chaudhary, Wenzek, Guzm{\'a}n, Grave, Ott, Zettlemoyer, and Stoyanov]{conneau2020xlmr}
Alexis Conneau, Kartikay Khandelwal, Naman Goyal, Vishrav Chaudhary, Guillaume Wenzek, Francisco Guzm{\'a}n, Edouard Grave, Myle Ott, Luke Zettlemoyer, and Veselin Stoyanov.
\newblock Unsupervised cross-lingual representation learning at scale.
\newblock In \emph{Proceedings of ACL}, 2020.

\bibitem[Dahl et~al.(2024)Dahl, Magesh, Suzgun, and Ho]{dahl2024large}
Matthew Dahl, Varun Magesh, Mirac Suzgun, and Daniel~E Ho.
\newblock Large legal fictions: Profiling legal hallucinations in large language models.
\newblock \emph{Journal of Legal Analysis}, 16\penalty0 (1):\penalty0 64--93, 2024.

\bibitem[Fowler et~al.(2007)Fowler, Johnson, Spriggs, Jeon, and Wahlbeck]{fowler2007network}
James~H Fowler, Timothy~R Johnson, James~F Spriggs, Sangick Jeon, and Paul~J Wahlbeck.
\newblock Network analysis and the law: Measuring the legal importance of precedents at the {U.S.} {Supreme Court}.
\newblock \emph{Political Analysis}, 15\penalty0 (3):\penalty0 324--346, 2007.

\bibitem[Guha et~al.(2023)Guha, Nyarko, Ho, R{\'e}, Chilton, Nanamori, Holzenberger, et~al.]{guha2023legalbench}
Neel Guha, Julian Nyarko, Daniel~E Ho, Christopher R{\'e}, Adam Chilton, Alex Nanamori, Nils Holzenberger, et~al.
\newblock {LegalBench}: A collaboratively built benchmark for measuring legal reasoning in large language models.
\newblock In \emph{NeurIPS Datasets and Benchmarks Track}, 2023.
\newblock arXiv:2308.11462.

\bibitem[Hu et~al.(2022)Hu, Shen, Wallis, Allen-Zhu, Li, Wang, Wang, and Chen]{hu2022lora}
Edward~J Hu, Yelong Shen, Phillip Wallis, Zeyuan Allen-Zhu, Yuanzhi Li, Shanan Wang, Lu~Wang, and Weizhu Chen.
\newblock {LoRA}: Low-rank adaptation of large language models.
\newblock In \emph{ICLR}, 2022.

\bibitem[Katz et~al.(2024)Katz, Bommarito, Gao, and Arredondo]{katz2024gpt4}
Daniel~Martin Katz, Michael~James Bommarito, Shang Gao, and Pablo Arredondo.
\newblock {GPT-4} passes the bar exam.
\newblock \emph{Philosophical Transactions of the Royal Society A}, 382\penalty0 (2270), 2024.

\bibitem[Magesh et~al.(2025)Magesh, Surani, Dahl, Suzgun, Manning, and Ho]{magesh2024hallucination}
Varun Magesh, Faiz Surani, Matthew Dahl, Mirac Suzgun, Christopher~D Manning, and Daniel~E Ho.
\newblock Hallucination-free? assessing the reliability of leading {AI} legal research tools.
\newblock \emph{Journal of Empirical Legal Studies}, 22:\penalty0 216--242, 2025.
\newblock arXiv:2405.20362.

\bibitem[Mazzega et~al.(2009)Mazzega, Bourcier, and Boulet]{mazzega2009network}
Pierre Mazzega, Dani{\`e}le Bourcier, and Romain Boulet.
\newblock The network of {French} legal codes.
\newblock \emph{Artificial Intelligence and Law}, 17\penalty0 (3), 2009.

\bibitem[Min et~al.(2023)Min, Krishna, Lyu, Lewis, Yih, Koh, Iyyer, Zettlemoyer, and Hajishirzi]{min2023factscore}
Sewon Min, Kalpesh Krishna, Xinxi Lyu, Mike Lewis, Wen-tau Yih, Pang~Wei Koh, Mohit Iyyer, Luke Zettlemoyer, and Hannaneh Hajishirzi.
\newblock {FActScore}: Fine-grained atomic evaluation of factual precision in long form text generation.
\newblock In \emph{Proceedings of EMNLP}, 2023.

\bibitem[Mones and Arvidsson(2021)]{mones2021emergence}
Enys Mones and Adam Arvidsson.
\newblock Emergence of hierarchy in networked endorsement dynamics.
\newblock \emph{Proceedings of the National Academy of Sciences}, 118\penalty0 (16), 2021.

\bibitem[Niklaus et~al.(2023)Niklaus, Matoshi, Rani, Gallucci, and Stuermer]{niklaus2023lextreme}
Joel Niklaus, Veton Matoshi, Pooja Rani, Andrea Gallucci, and Matthias Stuermer.
\newblock {LEXTREME}: A multi-lingual and multi-task benchmark for the legal domain.
\newblock In \emph{Findings of EMNLP}, 2023.

\bibitem[Ouyang et~al.(2022)Ouyang, Wu, Jiang, Almeida, Wainwright, Mishkin, Zhang, Agarwal, Slama, Ray, et~al.]{ouyang2022training}
Long Ouyang, Jeffrey Wu, Xu~Jiang, Diogo Almeida, Carroll~L Wainwright, Pamela Mishkin, Chong Zhang, Sandhini Agarwal, Katarina Slama, Alex Ray, et~al.
\newblock Training language models to follow instructions with human feedback.
\newblock In \emph{NeurIPS}, 2022.

\bibitem[Ovcharov(2026{\natexlab{a}})]{ovcharov2026citation}
Volodymyr Ovcharov.
\newblock Automatic construction of a legal citation graph from 100 million {Ukrainian} court decisions: Large-scale extraction, topological analysis, and ontology-driven clustering.
\newblock \emph{arXiv preprint arXiv:2605.15362}, 2026{\natexlab{a}}.

\bibitem[Ovcharov(2026{\natexlab{b}})]{ovcharov2026dissertation}
Volodymyr~V Ovcharov.
\newblock \emph{Methods for Ensuring Verifiability of Large Language Models in the Legal Domain}.
\newblock PhD thesis, National Academy of Sciences of Ukraine, 2026{\natexlab{b}}.

\bibitem[Rafailov et~al.(2023)Rafailov, Sharma, Mitchell, Manning, Ermon, and Finn]{rafailov2023direct}
Rafael Rafailov, Archit Sharma, Eric Mitchell, Christopher~D Manning, Stefano Ermon, and Chelsea Finn.
\newblock Direct preference optimization: Your language model is secretly a reward model.
\newblock In \emph{NeurIPS}, 2023.

\bibitem[Schimanski et~al.(2026)Schimanski, Ni, Kraus, and Leippold]{schimanski2026citeaudit}
Tobias Schimanski, Jingwei Ni, Mathias Kraus, and Markus Leippold.
\newblock {CiteAudit}: You cited it, but did you read it? {A} benchmark for verifying scientific references in the {LLM} era.
\newblock \emph{arXiv preprint arXiv:2602.23452}, 2026.

\bibitem[Weiser(2023)]{weiser2023mata}
Benjamin Weiser.
\newblock Here's what happens when your lawyer uses {ChatGPT}.
\newblock \emph{The New York Times}, May 2023.

\bibitem[Winkels et~al.(2011)Winkels, de~Ruyter, and Kroese]{winkels2011determining}
Radboud Winkels, Jelle de~Ruyter, and Henryk Kroese.
\newblock Determining authority of dutch case law.
\newblock In \emph{Proceedings of JURIX}, 2011.

\bibitem[Yang et~al.(2024)Yang, Yang, et~al.]{yang2024qwen25}
An~Yang, Baosong Yang, et~al.
\newblock {Qwen2.5} technical report.
\newblock \emph{arXiv preprint arXiv:2412.15115}, 2024.

\end{thebibliography}

\end{document}